\documentclass{article}

\usepackage[nonatbib, preprint]{neurips_2022}

\usepackage[utf8]{inputenc} % allow utf-8 input
\usepackage[T1]{fontenc}    % use 8-bit T1 fonts
\usepackage[dvipsnames]{xcolor}
\usepackage[pagebackref=true,breaklinks=true,letterpaper=true,colorlinks,bookmarks=false, citecolor=ForestGreen]{hyperref}     % hyperlinks
\usepackage{url}            % simple URL typesetting
\usepackage{booktabs}       % professional-quality tables
\usepackage{amsfonts}       % blackboard math symbols
\usepackage{nicefrac}       % compact symbols for 1/2, etc.
\usepackage{microtype}      % microtypography

\usepackage{graphicx}
\usepackage{comment}
\usepackage{amsmath,amssymb} % define this before the line numbering.
\usepackage{xspace}
\usepackage[font=footnotesize,labelfont=bf]{caption} 
\usepackage{caption}
\usepackage{mathtools}
\usepackage{graphicx}
\usepackage{soul}
\usepackage{xspace}
\usepackage{cleveref}
\usepackage{enumitem}

\usepackage[square,numbers,sort&compress]{natbib}
\usepackage{bbding}
\usepackage{wasysym}
\usepackage{amssymb}
\usepackage{float}
\usepackage{dblfloatfix}
\usepackage{footnote}
\usepackage{array} % For tables
\usepackage{dblfloatfix}
\usepackage{transparent}
\usepackage{xspace}
\usepackage{multirow}
\usepackage{amsfonts}
\usepackage{pifont}
\usepackage{algorithm}
\usepackage{listings}
\usepackage{bbold}
\usepackage{wrapfig}
\usepackage{subcaption}

\usepackage{etoolbox} % from moco paper
\makeatletter
\AfterEndEnvironment{algorithm}{\let\@algcomment\relax}
\AtEndEnvironment{algorithm}{\kern2pt\hrule\relax\vskip3pt\@algcomment}
\let\@algcomment\relax
\newcommand\algcomment[1]{\def\@algcomment{\footnotesize#1}}
\renewcommand\fs@ruled{\def\@fs@cfont{\bfseries}\let\@fs@capt\floatc@ruled
  \def\@fs@pre{\hrule height.8pt depth0pt \kern2pt}%
  \def\@fs@post{}%
  \def\@fs@mid{\kern2pt\hrule\kern2pt}%
  \let\@fs@iftopcapt\iftrue}
\makeatother

\def\etal{\emph{et~al.}}
\def\eg{\emph{e.g.}}
\def\ie{\emph{i.e.}}
\def\etc{\emph{etc}}
\DeclareMathOperator*{\argmax}{arg\,max}

\newcolumntype{x}[1]{>{\centering\arraybackslash}p{#1pt}}
\newcolumntype{y}[1]{>{\raggedright\arraybackslash}p{#1pt}}
\newcolumntype{z}[1]{>{\raggedleft\arraybackslash}p{#1pt}}
\newlength\savewidth\newcommand\shline{\noalign{\global\savewidth\arrayrulewidth  \global\arrayrulewidth 1pt}\hline\noalign{\global\arrayrulewidth\savewidth}}
\newcommand{\tablestyle}[2]{\setlength{\tabcolsep}{#1}\renewcommand{\arraystretch}{#2}\centering\footnotesize}

\definecolor{Highlight}{HTML}{39b54a}  % green
\definecolor{green}{HTML}{39b54a} % more transparent
\definecolor{red}{HTML}{cb4335} % red

\DeclareCaptionLabelFormat{andtable}{#1~#2  \&  \tablename~\thetable}

\title{Discovering Object Masks with Transformers for \\ Unsupervised Semantic Segmentation}

\author{
Wouter Van Gansbeke$^{1}$ \quad Simon Vandenhende$^{1}$ \quad Luc Van Gool$^{1,2}$  \\
\\
\normalfont $^1$ KU Leuven/ESAT-PSI  \quad \normalfont $^2$ ETH Zurich/CVL}

\begin{document}

\maketitle

\begin{abstract}
The task of unsupervised semantic segmentation aims to cluster pixels into semantically meaningful groups. Specifically, pixels assigned to the same cluster should share high-level semantic properties like their object or part category. This paper presents MaskDistill: a novel framework for unsupervised semantic segmentation based on three key ideas. First, we advocate a \emph{data-driven} strategy to generate object masks that serve as a pixel grouping prior for semantic segmentation. This approach omits handcrafted priors, which are often designed for specific scene compositions and limit the applicability of competing frameworks. Second, MaskDistill clusters the object masks to obtain pseudo-ground-truth for training an initial object segmentation model. Third, we leverage this model to filter out low-quality object masks. This strategy mitigates the noise in our pixel grouping prior and results in a clean collection of masks which we use to train a final segmentation model. By combining these components, we can considerably outperform previous works for unsupervised semantic segmentation on PASCAL ($+11$\% mIoU) and COCO ($+4$\% mask AP$_{50}$). Interestingly, as opposed to existing approaches, our framework does not latch onto low-level image cues and is not limited to object-centric datasets. The code and models are available.~\footnote{Code: \url{https://github.com/wvangansbeke/MaskDistill}}
\end{abstract}

\section{Introduction}
\label{sec: introduction}
The task of assigning a class label to each pixel in an image -- known as \textit{semantic segmentation} -- has been researched extensively~\cite{long2015fully,minaee2021image}. Semantic segmentation tools are used in many domains like autonomous driving~\cite{cordts2016cityscapes}, medical imaging~\cite{menze2014multimodal}, and agriculture~\cite{chiu2020agriculture}. Today, researchers tackle the segmentation task via deep convolutional nets~\cite{he2016deep} which learn hierarchical image representations from fully-annotated datasets~\cite{everingham2010pascal,lin2014microsoft} where each pixel is associated with a category label. However, collecting such annotations consumes large amounts of time and money~\cite{bearman2016s}. Therefore, several works explored less labor-intensive forms of annotations to train a segmentation model, \eg, scribbles~\cite{lin2016scribblesup,tang2018normalized,tang2018regularized}, bounding boxes~\cite{dai2015boxsup,khoreva2017simple,papandreou2015weakly}, clicks~\cite{bearman2016s}, and image-level tags~\cite{papandreou2015weakly,tang2018regularized,xu2015learning}. Others studied semi-supervised methods~\cite{dai2015boxsup,hong2015decoupled,hung2018adversarial,papandreou2015weakly} that improve the performance by leveraging additional unlabeled images during training. In this paper, we go a step further and learn a segmentation model in a self-supervised way. Specifically, the goal is to learn a clustering function that assigns semantically related pixels to the same cluster without relying on human labeling.

To realize this concept, \emph{end-to-end} methods~\cite{xu2019invariant,caron2018deep,ouali2020autoregressive} learned a clustering function by imposing consistency on the cluster assignments of pixels in augmented views of an image. However, these methods tend to latch onto low-level image cues like color or texture (see~\cite{van2020scan,cho2021picie}). In particular, the clusters strongly depend on the network's initialization leading to degenerate solutions. Unlike these methods, we do not adopt an end-to-end strategy but follow the works discussed next.

Another group of works proposed a \emph{bottom-up} approach for tackling the problem. First, they leverage a low- or mid-level visual prior like edge detection~\cite{hwang2019segsort,zhang2020self} or saliency estimation~\cite{van2021unsupervised} to find image regions that likely share the same semantics. In a second step, they use the image regions to learn pixel-embeddings that capture semantic information. In particular, the image regions serve as a regularizer which removes the segmentation's dependence on the network initialization. The pixel-embeddings can subsequently be clustered via K-means to obtain an image segmentation. While bottom-up approaches report better results, they suffer from several drawbacks too. Most importantly, their dependence on a handcrafted prior, \eg, edges or saliency, to group pixels limits their usage. For example, saliency estimation only applies to object-centric images. Additionally, several works require annotations to identify the appropriate image regions. For example, Hwang~\etal~\cite{hwang2019segsort} use boundary annotations from~\cite{martin2001database}.

This paper presents \textbf{MaskDistill}, a novel framework that addresses the above limitations. Like bottom-up methods, MaskDistill first identifies groups of pixels that likely belong to the same object. Since objectness is a high-level construct~\cite{kuo2015deepbox}, we avoid using a handcrafted prior and instead advocate a data-driven approach. We observe that self-supervised vision transformers~\cite{caron2021emerging,chen2021empirical} learn spatially structured image representations. In particular, it's possible to distill highly accurate object masks through the attention layers in vision transformers~\cite{dosovitskiy2020image,vaswani2017attention}.  Different from existing works~\cite{van2021unsupervised,hwang2019segsort,zhang2020self} which rely on handcrafted priors, this facilitates the scaling of our framework to more challenging datasets. In particular, handcrafted priors tend to be designed for specific scene compositions. For example, saliency estimation works well for images with few objects (\eg, PASCAL~\cite{everingham2010pascal}) but fails for more complex scenes (\eg, COCO~\cite{lin2014microsoft}). Our framework does not suffer from this problem (see Section~\ref{sec: experiments}).

We cluster the object masks and use the result as pseudo-ground-truth to train an object segmentation model, \eg, Mask R-CNN~\cite{he2017mask}. As discussed in Section~\ref{sec: sem_seg}, this model predicts object mask candidates together with their confidence scores. We empirically observed that higher confidence scores correlate with object masks of better quality (see Figure~\ref{fig: conf_threshold}). Based upon this observation, we construct a cleaner set of object masks by leveraging the model's predictions. In particular, we filter out predictions with low confidence scores for each image. The resulting set of object masks is used as pseudo-ground-truth to train a final semantic segmentation model. 

In summary, our contributions are: \textbf{(i)} we develop a novel bottom-up framework to tackle the task of unsupervised semantic segmentation (Section~\ref{sec: method}), \textbf{(ii)} we present a data-driven strategy to get a pixel grouping prior for semantic segmentation based on self-supervised transformer models (Section~\ref{sec: mask_distillation}), \textbf{(iii)} we analyze the use of confident object mask candidates to refine the segmentation results (Section~\ref{sec: sem_seg}), and \textbf{(iv)} we obtain state-of-the-art results on the well-known PASCAL~\cite{everingham2010pascal} and COCO~\cite{lin2014microsoft} datasets under the unsupervised setup (Section~\ref{sec: experiments}).
\section{Related Work}
\label{sec: related_work}
\paragraph{Unsupervised Semantic Segmentation.} Several works tried to segment \emph{stuff} categories, \eg, sky, grass, mountain,~\etc. For example,~\cite{xu2019invariant,ouali2020autoregressive} maximized the mutual information between augmented views to learn a segmentation model. Others~\cite{caron2018deep,cho2021picie} iteratively refined the segmentation model's features via a clustering objective. However, these methods rely on the architectural prior which makes them prone to degenerate solutions, and limits their use to small-scale problems, \eg, segmenting roads and vegetation in satellite imagery. We refer to~\cite{van2020scan,van2021unsupervised,cho2021picie} for an analysis. This paper differs from these works in two ways. First, we segment object rather than stuff categories. This setting aligns better with popular segmentation benchmarks, \eg, PASCAL~\cite{everingham2010pascal}, COCO~\cite{lin2014microsoft}, \etc. Furthermore, learning object-centric representations is a key component of machine intelligence with applications in augmented reality~\cite{abu2018augmented,grauman2021ego4d}. Second, unlike the referred works, we do not employ an end-to-end learning strategy which makes our framework less dependent on the architectural prior.

As mentioned, we focus on segmenting \emph{object} categories. Earlier works~\cite{van2021revisiting,hwang2019segsort,zhang2020self} that studied this problem applied a two-step strategy. First, a handcrafted prior -- \eg, superpixels, boundary maps, or saliency -- is used to find groups of pixels that likely belong to the same object or part. Next, a pixel-level representation is learned that is discriminative of these groups. This allows the representations to be clustered via K-means to get an image segmentation. Our framework differs from these works in three ways. First, we do not use a handcrafted prior. Instead, we leverage the attention mechanism from self-supervised vision transformers~\cite{caron2021emerging,chen2021empirical} to mine object masks as a pixel grouping prior. This data-driven strategy makes fewer assumptions on the scene compositions which eases scaling of our approach (see Section~\ref{sec: learning_objectness}). Second, unlike earlier works~\cite{hwang2019segsort,zhang2020self,van2021unsupervised}, we do not use additional annotations, \eg, boundary maps, to construct our prior. Third, we directly predict the cluster assignments and avoid using K-means as post-processing. For completeness, we include a recent work~\cite{hamilton2022unsupervised} which also advocates a data-driven approach but considers fewer categories.

\paragraph{Unsupervised Object Detection.} The task of unsupervised object detection aims to produce object candidates without using human annotations. This concept is realized via a class-agnostic objectness scoring function~\cite{alexe2012measuring} which estimates the probability for an image window to contain an object. Existing methods learned such a function via foreground-background masks\cite{endres2014category,carreira2012cpmc}, superpixels~\cite{uijlings2013selective,manen2013prime} or edge information~\cite{zitnick2014edge}. Recent approaches~\cite{vo2021largescale,vo2020towards,simeoni2021localizing} have shown promising results on large-scale benchmarks, \eg, COCO~\cite{lin2014microsoft} and OpenImages~\cite{kuznetsova2020open}. In this work, we employ an object mask distillation strategy that is related to LOST~\cite{simeoni2021localizing}. However, unlike our method, LOST fails to generate multiple object mask candidates per image -- which is critical for the task of semantic segmentation. A few methods~\cite{greff2019multi,engelcke2019genesis,burgess2019monet} do produce several masks per image, but these are limited to small-scale problems (\eg, CLEVR~\cite{johnson2017clevr}).

\paragraph{Self-Supervised Representation Learning.} These works learn visual representations from unlabeled images by solving \emph{pretext tasks}. Some examples include predicting transformations~\cite{gidaris2018unsupervised,zhang2019aet}, predicting optical flow~\cite{mahendran2018cross,zhan2019self}, solving jigsaw puzzles~\cite{noroozi2016unsupervised,noroozi2018boosting}, predicting noise~\cite{bojanowski2017unsupervised}, performing clustering~\cite{asano20self,asano2020labelling,caron2018deep,yan2020clusterfit,caron2019unsupervised}, image colorization~\cite{iizuka2016let,larsson2017colorization,zhang2016colorful}, inpainting~\cite{pathak2016context}, predictive coding~\cite{oord2018representation} \etc. The instance discrimination task~\cite{wu2018unsupervised,chen2020simple,he2019momentum,oord2018representation,chen2021empirical} and its alternatives~\cite{grill2020bootstrap,caron2021emerging,chen2020exploring} outperform their supervised counterparts when transferring the representations to various downstream tasks, \eg, object detection. In this work, we explore self-supervised learning to capture objectness, allowing us to mine object mask candidates in a data-driven way.
\section{Method}
\label{sec: method}
\begin{figure}
    \centering
    \includegraphics[width=1.0\linewidth]{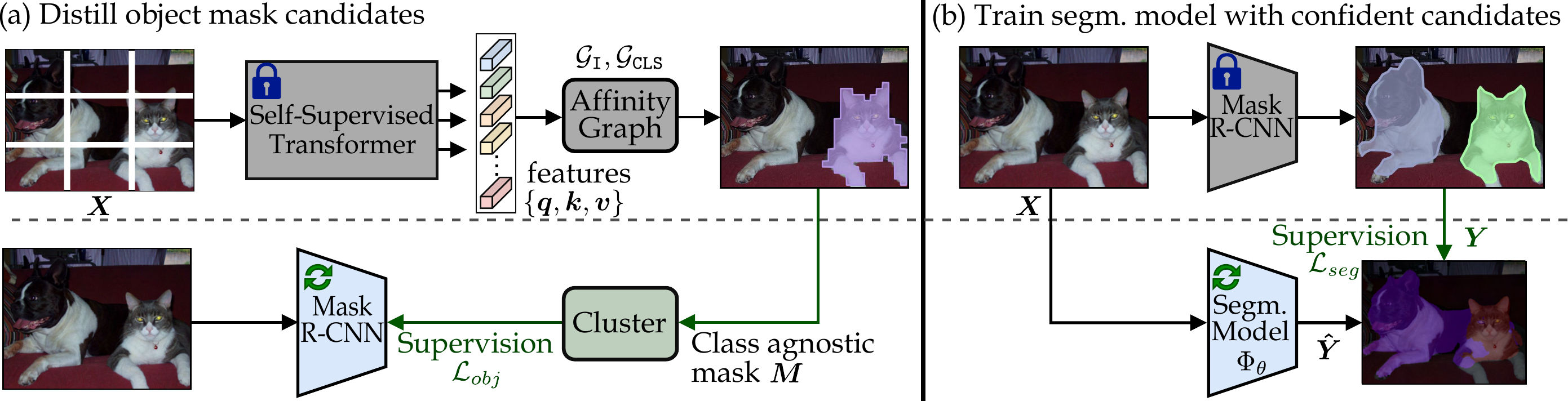}
    \caption{\textbf{Overview of MaskDistill.} We present a simple framework for unsupervised semantic segmentation. \textbf{(Left)} We distill class-agnostic object masks from self-supervised transformers~\cite{dosovitskiy2020image}. For each image, we commit to the most discriminative object. The found masks are subsequently clustered and used as pseudo-ground-truth to train an object segmentation model, \ie, Mask R-CNN~\cite{he2017mask} (Section~\ref{sec: mask_distillation}). \textbf{(Right)} The learned Mask R-CNN model predicts multiple object mask candidates per image with their respective confidence scores. We use the most confident predictions as pseudo-ground-truth to train a segmentation model (Section~\ref{sec: sem_seg}).}
    \label{fig: overview}
    \vspace{-1.4em}
\end{figure}

Our approach follows a bottom-up scheme to tackle the unsupervised semantic segmentation task. First, we advocate a data-driven approach to mine object masks via self-supervised vision transformers (Section~\ref{sec: learning_objectness}). Second, we distill multiple object masks per image via an object segmentation model, \ie, Mask R-CNN (Section~\ref{sec: mask_distillation}). Third, we discuss how to train a final segmentation model using the found object masks (Section~\ref{sec: sem_seg}). As a key component, we use only object masks with high confidence scores. This strategy mitigates the noise introduced during the mask distillation step. Figure~\ref{fig: overview} shows an overview of our proposed MaskDistill framework.

\subsection{Learning Objectness}
\label{sec: learning_objectness}
\begin{wrapfigure}[15]{r}{0.45\textwidth}
    \centering
    \vspace{-1.1em}
    \includegraphics[width=1.0\linewidth]{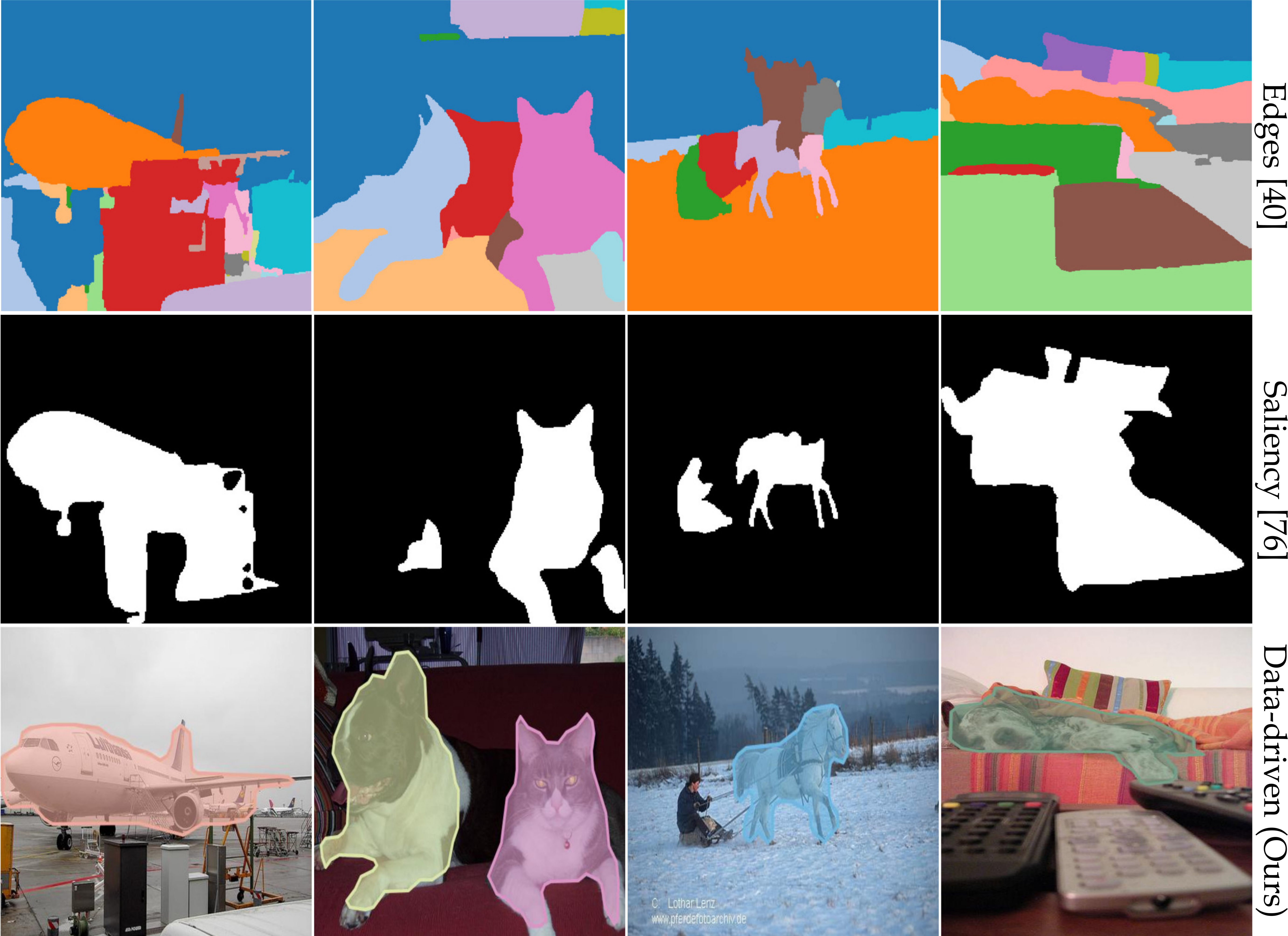}
    \captionof{figure}{\textbf{Pixel grouping strategies.} Our masks (bottom) capture high-level object information.} 
    \label{fig: priors_examples}
\end{wrapfigure}
End-to-end approaches~\cite{xu2019invariant,caron2018deep,ouali2020autoregressive} are unlikely to discover image regions that pertain to high-level object categories~\cite{van2020scan}, \eg, birds, cats, buildings, \etc. For this reason, we follow prior work~\cite{hwang2019segsort,van2021unsupervised,uijlings2013selective}, and advocate a bottom-up approach to tackle the task of unsupervised semantic segmentation. In particular, it's advantageous to break down an image into its different components first, before going after its semantic segmentation. Existing methods achieve this via a handcrafted low-level (\eg, superpixels or edges) or mid-level (\eg, saliency) pixel grouping prior. However, such priors are suboptimal. A low-level prior based on superpixels or edges produces an over-segmentation of the image, which yields image regions with low semantic content (see the top row in Figure~\ref{fig: priors_examples}). Differently, a mid-level prior can aggregate parts from different objects (see the middle row in Figure~\ref{fig: priors_examples}). To address these drawbacks, we propose to obtain a pixel grouping prior in a data-driven way by relying on self-supervised representation learning. The bottom row in Figure~\ref{fig: priors_examples} shows some examples. Unlike handcrafted pixel grouping priors, our approach generates object masks that align with \textit{true} objects. For example, we correctly identify the entire plane in the 1st column, while other methods fail to do so.

In this paper, we build on self-supervised vision transformers~\cite{caron2021emerging,chen2021empirical} to mine object masks. The reason for this decision is three-fold. First, transformers reason at patch-level~\cite{vaswani2017attention,dosovitskiy2020image} which allows us to construct an affinity graph expressing the similarity between different image regions. Second, self-supervised vision transformers learn rich spatial representations that capture object information~\cite{caron2021emerging,van2021revisiting} which facilitates their use for mining object masks. Moreover, the representations encode detailed information about each image component which can exceed a human-defined taxonomy. Third, self-supervised vision transformers do not rely on human annotations which allows us to take advantage of large unlabeled datasets~\cite{goyal2019scaling}. Motivated by these findings, we propose to distill object information from the final self-attention layer in the vision transformer~\cite{dosovitskiy2020image}.

% Overview Figure (Method)

\subsection{Distilling Object Masks Using Self-Attention}
\label{sec: mask_distillation}
\paragraph{Preliminaries.} We reshape an image $\boldsymbol{X} \in \mathbb{R}^{ H\times W \times 3}$ into a sequence of $N$ patches. Each patch is of size $S \times S$ pixels. We refer to the image patches as \emph{patch tokens} [\texttt{I}]. The patch tokens are further concatenated with a special \emph{classification token} [\texttt{CLS}] resulting in the input sequence $\boldsymbol{X'}$ which consists of $N+1$ tokens. We use the features $\{\boldsymbol{q}(h), \boldsymbol{k}(h)\}$ from the final multihead self-attention (MSA) block to compute object masks, where each head $h$ performs a single self-attention operation. The self-supervised vision transformer is initialized with weights from~\cite{caron2021emerging}. We refer the interested reader to~\cite{dosovitskiy2020image,vaswani2017attention} for more information on the self-attention mechanism in transformers. 

\paragraph{Construct Affinity Graphs.} Following prior work~\cite{caron2021emerging,simeoni2021localizing,vaswani2017attention}, we construct an affinity graph to measure the similarity between image patches. Given the input sequence $\boldsymbol{X}'$, we compute the affinity vector $\boldsymbol{a}_\texttt{CLS}$ as the pairwise similarities between the classification token \texttt{[CLS]} and the patch tokens \texttt{[I]} in the final MSA block. Similarly, the affinity matrix $\boldsymbol{A}_\texttt{I}$ measures the pairwise similarities between all pairs of patch tokens \texttt{[I]}. In particular, the element $\boldsymbol{A}^{ij}$ is computed between two tokens of the sequence, $i$ and $j$, as the dot product of their feature representations, $\boldsymbol{f}^i$ and $\boldsymbol{f}^j$ where $\boldsymbol{f} \in \{\boldsymbol{q}(h), \boldsymbol{k}(h)\}$. Finally, we average the affinities over the attention heads $\mathcal{H}$. The edges $\mathcal{E}_\texttt{CLS}$ and $\mathcal{E}_\texttt{I}$ in the graphs, $\mathcal{G}_{\texttt{CLS}}$ and $\mathcal{G}_{\texttt{I}}$, are defined by their associated affinity weights, shown in Eq.~\ref{eq: attn_sum} and~\ref{eq: features_sum}: 
\begin{align}
    \boldsymbol{a}_\texttt{CLS} &= \frac{1}{|\mathcal{H}|}\sum_{h \in \mathcal{H}} \boldsymbol{q}_\texttt{CLS}(h)\cdot \boldsymbol{k}^{\top}_\texttt{I}(h)  & \boldsymbol{a}_{\texttt{CLS}} &\in \mathbb{R}^{1 \times N}, \label{eq: attn_sum} \\
    \boldsymbol{A}_\texttt{I} &= \frac{1}{|\mathcal{H}|}\sum_{h \in \mathcal{H}} \boldsymbol{k}_\texttt{I}(h)\cdot \boldsymbol{k}^{\top}_\texttt{I}(h) & \boldsymbol{A}_\texttt{I} &\in \mathbb{R}^{N \times N}. \label{eq: features_sum}
\end{align}

\paragraph{Select Discriminative Tokens.} Our goal is to select patch tokens that likely correspond to an object part. In particular, we focus on the top-$k$ responses according to the affinities w.r.t. the [\texttt{CLS}] token $\boldsymbol{a}_\texttt{CLS}$. Formally, we define the set of patches $\mathcal{P}=\{j~|~\mathcal{E}(\texttt{CLS},j)~\text{is top-}k \in \mathcal{G}_\texttt{CLS}\}$, where $\mathcal{E}(\texttt{CLS}, j)$ denotes the directed edge from the classification token [\texttt{CLS}] to a patch token [\texttt{I}$^j$] in graph $\mathcal{G}_\texttt{CLS}$. In addition, we define the patch with the largest (\textit{i.e.,} top-1) affinity in $\boldsymbol{a}_\texttt{CLS}$ as the \textit{source} patch $s = {\argmax_j \boldsymbol{a}^j_\texttt{CLS}}$.  This region tends to correspond to the most discriminative image component, \eg, the beak of a bird, the horn of a rhino,~\etc. 

\paragraph{Construct Initial Masks.} We generate a \textit{single} object mask $\boldsymbol{M}_{s}\in \{0,1\}^{1\times N}$ per image $\boldsymbol{X}$ based on its source $s$ and proposals $\mathcal{P}$. The source $s$ should belong to the predicted object mask as it represents the object's most discriminative part. We follow~\cite{simeoni2021localizing} to diffuse the information from $s$ to the proposals $\mathcal{P}$.  In particular, only patches in $\mathcal{P}$ that are similar to $s$ are further considered as proposals $\mathcal{P}'=\{j~|~j\in\mathcal{P} \wedge \boldsymbol{A}^{sj}_\texttt{I}>0\}$. The object mask $\boldsymbol{M}_{s}$ is set to 1 at location $j$ only if $\sum_{i \in \mathcal{P}'}\boldsymbol{A}^{ij}_{\texttt{I}} > 0$. Consequently, patch $j$ belongs to the same object as $s$, if the total sum of pairwise similarities between $s$ and $\mathcal{P}'$ is positive. Finally, the obtained mask is reshaped and upsampled to the original image size $(H, W)$ using nearest neighbor interpolation, resulting in $\boldsymbol{M} \in \{0,1\}^{H\times W}$.  
\paragraph{Distill Mask R-CNN.} To produce \textit{multiple} object mask candidates per image, we train a region proposal network, \ie, Mask R-CNN~\cite{he2017mask}. This object segmentation model requires the class $c$, the bounding box coordinates $b$, and the foreground-background mask $\boldsymbol{M}$ for each image. Notice that we obtained the object masks and their corresponding bounding box coordinates in the previous step. However, these masks are class-agnostic. In order to assign a class label $c$ to each mask, we apply a clustering algorithm (\eg, K-means~\cite{lloyd1982least}) to the output [\texttt{CLS}] tokens of the masked images. Now, we can train Mask R-CNN via the following objective function:
\begin{equation}
    \mathcal{L}_{obj} = 
    \mathcal{L}_{class}(\hat{c}, c) +
    \mathcal{L}_{bbox}(\hat{b}, b) + 
     \mathcal{L}_{mask}(\hat{\boldsymbol{M}}, \boldsymbol{M}),
\label{eq: mask_rcnn_loss}
\end{equation}
where $\hat{c}, \hat{b}$ and $\hat{\boldsymbol{M}}$ denote the predicted class, bounding box and mask. Importantly, the trained model predicts multiple object mask candidates per image with their associated confidence scores. We leverage these predictions as pseudo-ground-truth to train a segmentation model in the next section. 

\paragraph{Discussion.} Like prior work~\cite{caron2021emerging,simeoni2021localizing}, MaskDistill constructs an affinity graph from the query and key features, respectively $\boldsymbol{q}(h)$ and $\boldsymbol{k}(h)$, in the final MSA block to produce an object mask. However, our method differs in two important components. First, we can generate \textit{multiple} candidates for each image. This is crucial when tackling scene-centric datasets. Second, we use the top-$k$ affinities in $\mathcal{G}_{\texttt{CLS}}$ to generate the patch proposals $\mathcal{P}$ and their initial object mask $M$. Notice that this strategy does not make assumptions about the underlying scene composition as in~\cite{simeoni2021localizing,van2021unsupervised}, \eg, the object should be salient or enclose a smaller area than the background. We empirically observe that our proposed approach results in better performance (see Section~\ref{sec: sota}).

\subsection{Training a Segmentation Model from Noisy Object Mask Candidates}
\label{sec: sem_seg}
Consider the set of images $\mathcal{X} = \{\boldsymbol{X}_1, \hdots, \boldsymbol{X}_{|\mathcal{D}|}\}$ with their corresponding object mask candidates $\mathcal{M} = \{\boldsymbol{M}_1, \hdots, \boldsymbol{M}_K \}$ and confidence scores $\mathcal{S} = \{s_1, \hdots, s_K \}$ -- obtained via the Mask R-CNN model from Section~\ref{sec: mask_distillation}.\footnote{$K$ is typically much larger than the number of images $|\mathcal{D}|$ in the dataset as Mask R-CNN returns multiple object mask candidates per image.} Some of the masks will inevitably get assigned to the wrong cluster or won't align with an object or part. Interestingly, we experimentally observe that masks for which the model is very confident ($s_i \approx 1$) tend to be correct (see experiment in Section~\ref{sec: ablation_studies}). Unlike previous methods~\cite{van2021unsupervised,hwang2019segsort,zhang2020self}, this allows us to leverage confidence scores to suppress the influence of the noise in our prior. Specifically, we only accept confident predictions from Mask R-CNN via a threshold $\tau$ as $\{\boldsymbol{M}_i | s_i \in \mathcal{S} \wedge s_i > \tau \}$. Finally, we aggregate the masks belonging to the same image to obtain an initial semantic segmentation per image $\mathcal{Y} = \{\boldsymbol{Y}_1, \hdots, \boldsymbol{Y}_{|\mathcal{D}|}\}$. We only keep the most confident mask when two candidates overlap. The constructed masks serve as pseudo-ground-truth to train a semantic segmentation model.

Finally, we train a semantic segmentation model $\Phi_\theta: \mathbb{R}^{H\times W \times 3} \rightarrow \mathbb{R}^{H\times W \times C}$ parameterized with weights $\theta$. This function terminates in a softmax operation to perform a soft assignment over the clusters $\mathcal{C} = \{1, \ldots, C \}$. To overcome class imbalance while simultaneously obtaining fine-grained segmentation results, we adopt a hard pixel mining strategy based on~\cite{wu2016bridging}. The top-$k$ most difficult pixels $\mathcal{T}$ are selected in each batch to train $\Phi_\theta$. In particular, the objective function becomes:
\begin{equation}
    \mathcal{L}_{seg} = -\frac{1}{|\mathcal{T}|\cdot|\mathcal{C}|}\sum_{i\in\mathcal{T}}\sum_{c \in \mathcal{C}} \boldsymbol{Y}(i, c)\log\boldsymbol{\hat{Y}}(i, c),
\label{eq: ce_loss}
\end{equation}
where the obtained segmentation mask $\boldsymbol{Y}(i, c)$ is 1 if pixel $i$ belongs to class $c$ and 0 otherwise.
\section{Experiments}
\label{sec: experiments}
%%%%%%%%%%%%%%%%%%%%% NEW COMMANDS %%%%%%%%%%%%%%%%%%%
\definecolor{darkF7E0D5}{RGB}{0,0,0}
\newcommand{\rownumber}[1]{\textcolor{darkF7E0D5}{#1}}
\newcommand{\demph}[1]{\textcolor{Gray}{#1}}
\newcommand{\std}[1]{{\fontsize{5pt}{1em}\selectfont ~~$_\pm$$_{\text{#1}}$}}

\renewcommand{\hl}[1]{\textcolor{Highlight}{#1}}

\newcommand{\res}[3]{
\tablestyle{1pt}{1}
\begin{tabular}{z{16}y{18}}
{#1} &
\fontsize{7.0pt}{1em}\selectfont{~(${#2}${#3})}
\end{tabular}}

\newcommand{\reshl}[3]{
\tablestyle{1pt}{1} 
\begin{tabular}{z{16}y{18}}
{\textbf{#1}} &
\fontsize{7.0pt}{1em}\selectfont{~\hl{({${#2}$}\textbf{#3})}}
\end{tabular}}

\newcommand{\resrand}[2]{\tablestyle{1pt}{1} \begin{tabular}{z{16}y{18}} \demph{#1} & {} \end{tabular}}
\newcommand{\ressup}[2]{\tablestyle{1pt}{1} 
\begin{tabular}{z{16}y{18}} {#1} & {} \end{tabular}}
%%%%%%%%%%%%%%%%%%%%%%%%%%%%%%%%%%%%%%%%%%%%%%%%%%%%%%

\subsection{Experimental Setup}
\label{sec: experimental_setup}
\paragraph{Datasets.} We conduct experiments on two popular benchmarks: PASCAL~\cite{everingham2010pascal} and COCO~\cite{lin2014microsoft}. We follow prior work~\cite{hwang2019segsort,van2021unsupervised} and report our results on the 21 classes of PASCAL. The \texttt{train\_aug} and \texttt{val} splits are used for training and evaluation respectively. We use all 80 object categories on COCO -- a considerably challenging setting that is usually not considered for unsupervised semantic segmentation. We follow~\cite{simeoni2021localizing,vo2021largescale} and use the COCO20k subset for training and testing. This subset was introduced in~\cite{vo2020towards} to evaluate object detection methods for scene-centric images. During K-means clustering, we use all COCO images to improve the clustering performance.

\paragraph{Mask Distillation Setup.} We use the ViT-S~\cite{dosovitskiy2020image} vision transformer with a patch size of $16 \times 16$ pixels for constructing the affinity graphs. The weights are initialized via self-supervised pre-training on ImageNet~\cite{caron2021emerging}. We select the top-40\% most discriminative patch tokens in the graph $\mathcal{G}_\texttt{CLS}$ after resizing the smallest image side to $640$ pixels. In order to assign a category label to each mask, we apply K-means~\cite{lloyd1982least} on the output [\texttt{CLS}] tokens when using masked images as input. The Mask R-CNN model consists of a ResNet-50-C4 backbone. The training setup follows~\cite{he2019momentum}.

\paragraph{Segmentation Training Setup.} We use a DeepLab-v3~\cite{chen2017rethinking} segmentation model with dilated~\cite{yu2015multi} ResNet-50 backbone~\cite{he2016deep} to facilitate a fair comparison with~\cite{van2021unsupervised}. The weights are initialized via self-supervised MoCo~\cite{chen2021empirical} pre-training on ImageNet. We train the segmentation model for $45$ epochs using batches of size $16$. The weights are updated through SGD with momentum $0.9$ and weight decay $10^{-4}$. The learning rate is $2\cdot 10^{-3}$ at the start and reduced to $2\cdot 10^{-4}$ after 40 epochs. Further, we use confidence threshold $\tau = 0.9$ to select the most confident masks from our Mask R-CNN model (see Section~\ref{sec: sem_seg}). We keep the mask with the largest confidence score when thresholding excludes all predictions in an image from being used. The cross-entropy loss in Eq.~\ref{eq: ce_loss} uses the top-20\% hardest pixels. Following~\cite{van2021unsupervised}, we freeze the first two ResNet blocks to increase speed. 

\paragraph{Evaluation Protocols.} We benchmark our approach via the evaluation protocols from~\cite{van2021unsupervised,xu2019invariant}. \textbf{(i) Linear classifier:} We train a $1 \times 1$ convolutional layer on top of frozen features to predict the semantic classes. If the pixels are disentangled according to their semantic category, we should be able to solve the segmentation task via a low-capacity (linear) classifier. \textbf{(ii) Clustering:} We directly evaluate the quality of our clusters by comparing our predictions against the ground-truth annotations via Hungarian matching~\cite{kuhn1955hungarian}. The semantic segmentation results are evaluated via the mean IoU metric. 

\subsection{Ablation Studies}
\label{sec: ablation_studies}
\paragraph{Component Analysis.}
\begin{wraptable}[7]{r}{0.45\linewidth}
    \vspace{-1em}
    \tablestyle{4pt}{1.0}
    \caption{\textbf{Component analysis of MaskDistill.} Results on PASCAL for the clustering setup.}
    \resizebox{0.45\textwidth}{!}{
    \begin{tabular}{l|c|cc}
    \toprule
    \textbf{Setup}  & \textbf{Section} & \textbf{\texttt{val} mIoU} \\
    \hline
    Self-sup. vision transformer & \textcolor{red}{Sec.~\ref{sec: mask_distillation}} & 39.0 \\
    $+$ Mask R-CNN  & \textcolor{red}{Sec.~\ref{sec: mask_distillation}} & 42.0 \\
    $+$ Segmentation model  & \textcolor{red}{Sec.~\ref{sec: sem_seg}} &  45.8 \\
    \bottomrule
    \end{tabular}
    }
    \label{tab: ablation_baselines}
\end{wraptable}
Table~\ref{tab: ablation_baselines} analyzes the effect of different components of MaskDistill on the \texttt{val} set of PASCAL. We achieve $39.0\%$ mIoU (first row) when clustering the initial object masks via K-means. Recall that the object masks are obtained via the affinity graph $\mathcal{G}_\texttt{I}$ from a self-supervised vision transformer (Section~\ref{sec: mask_distillation}). The results are further improved when using predictions from a Mask R-CNN model trained with the initial object masks (from $39.0 \%$ to $42.0\%$ mIoU - second row). This shows that our object mask candidates capture high-level object information, which is hard to achieve through handcrafted priors. 
Finally, we capitalize on the confidence scores predicted by Mask R-CNN, and show additional gains with our training recipe from Section~\ref{sec: sem_seg}. In particular, by using only confident object mask candidates from Mask R-CNN, our segmentation results improve from $42.0\%$ to $45.8\%$ mIoU. For completeness, removing the hard pixel mining strategy, results in $45.5\%$ mIoU. We refer to the supplementary for additional ablation results. 

\begin{figure}[t]
\begin{minipage}[t]{.31\linewidth} % FIG 1
\centering
\includegraphics[width=\textwidth]{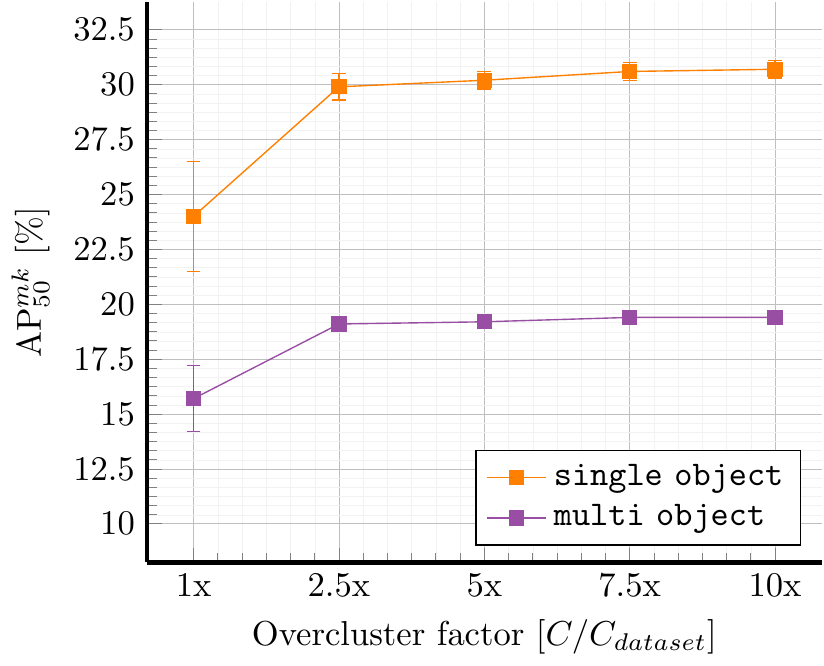}
\caption{Influence of the number of clusters in K-means.}
\label{fig: kmeans}
\end{minipage} %
\hspace{0.02\linewidth}
\begin{minipage}[t]{.31\linewidth} % FIG 2
\centering
\includegraphics[width=\textwidth]{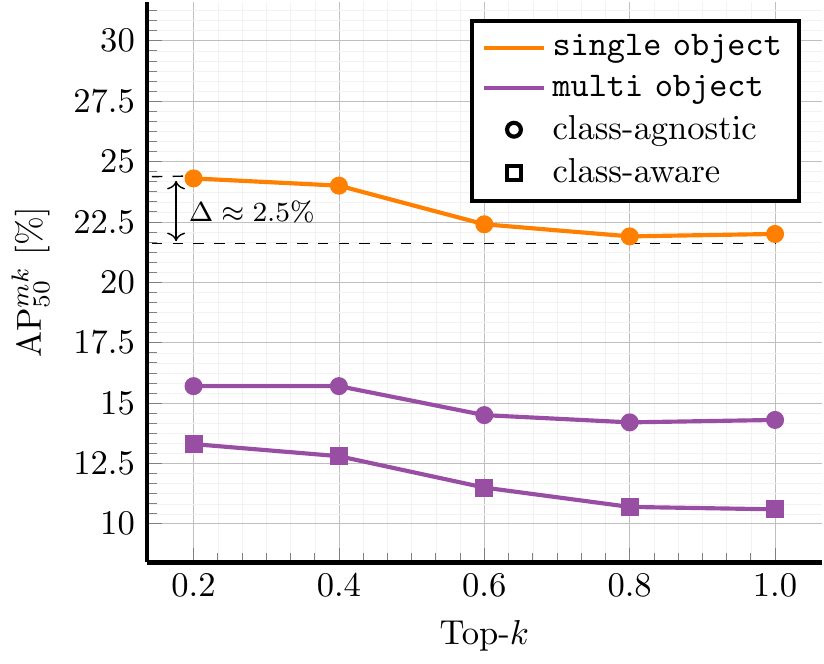}
\caption{Influence of the top-$k$ selection in $\mathcal{G_\texttt{CLS}}$.}
\label{fig: topk}
\end{minipage} %
\hspace{0.02\linewidth}
\begin{minipage}[t]{.31\linewidth} % FIG 3
\centering
\includegraphics[width=\textwidth]{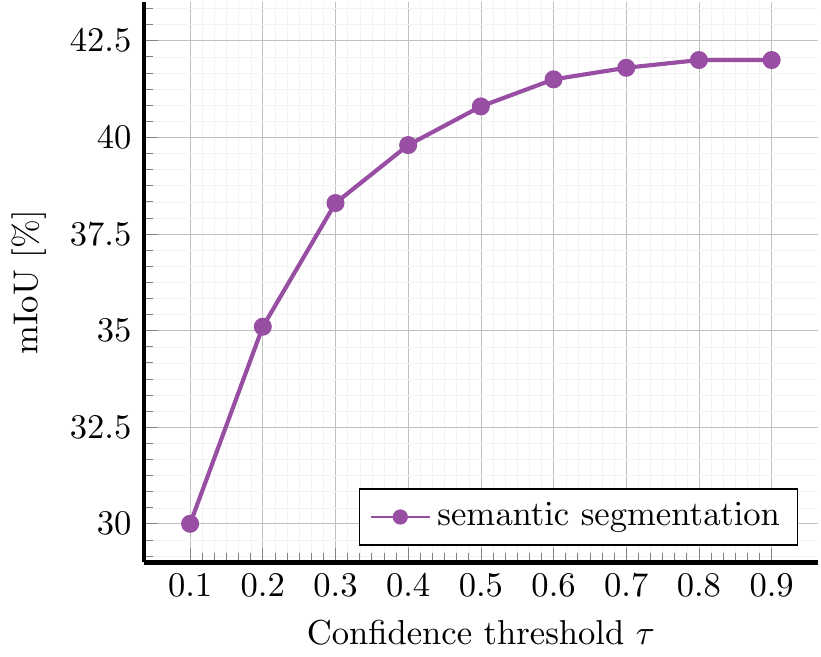}
\caption{Influence of the confidence threshold parameter.}
\label{fig: conf_threshold}
\end{minipage} %
\end{figure}

\paragraph{Hyperparameter analysis.} We study the influence of the hyperparameters on PASCAL and make the following observations: (\textbf{i}) Figure~\ref{fig: kmeans} quantifies the impact of changing the number of cluster $C$ during K-means clustering of the initial object masks (Section~\ref{sec: mask_distillation}). We adopt the overclustering procedure from~\cite{van2021unsupervised,van2020scan} and observe that the mask AP metric increases when we increase the amount of predicted clusters $C$. This means that the discovered clusters contain pixels of semantically related objects, irrespective of the amount $C$. (\textbf{ii}) Figure~\ref{fig: topk} shows the impact of the top-$k$ selection. In order to mitigate the influence of spurious details (\textit{e.g.,} background clutter), we select the top-$k$ patches in $\mathcal{G_\texttt{CLS}} $ which correspond to the most discriminative patch tokens (see Section~\ref{sec: mask_distillation}). To strike a good balance between the accuracy and the amount of proposals $|P|$, we set $k$ to $40\%$ in our experiments.
(\textbf{iii}) Figure~\ref{fig: conf_threshold} studies the influence of selecting the most confident object mask candidates with threshold $\tau$, discussed in Section~\ref{sec: sem_seg}. We observe that the mIoU score plateaus around $75\%$. Finally, these results show that our approach is not very sensitive to the used hyperparameters, \ie, number of clusters $C$, top-$k$ and threshold $\tau$. As a result, we use the same setup in all experiments (see Section~\ref{sec: experimental_setup}).

%%%% Figure part I %%%%

\subsection{Comparison to State-of-the-art}
\label{sec: sota}
\subsubsection{Semantic Segmentation}
\label{sec: sota_semseg}
\begin{figure}[t]
    \centering
    \includegraphics[width=1.0\linewidth,height=7.5cm]{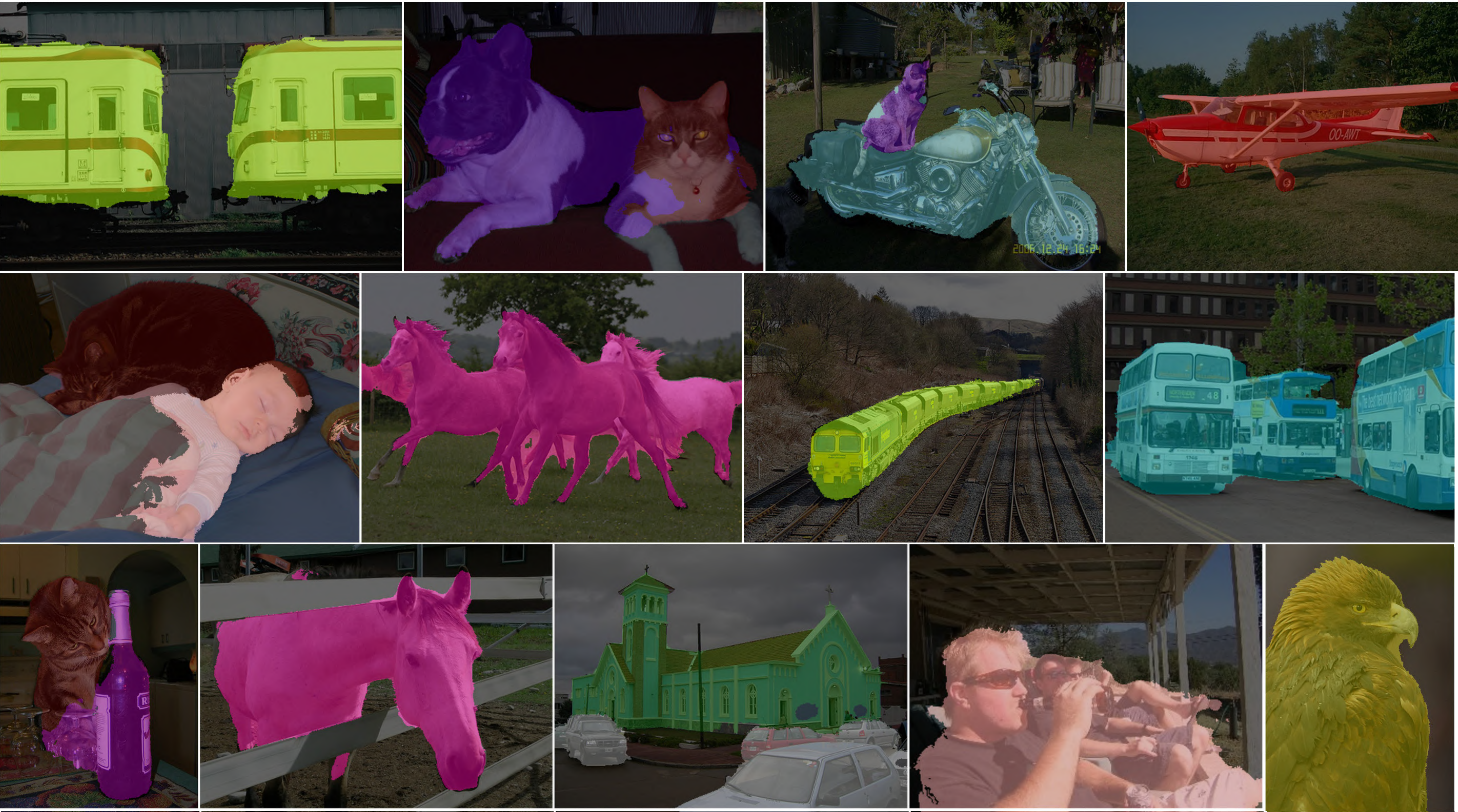}
    \caption{\textbf{Semantic segmentation results} of our method obtained under the clustering setup on PASCAL.}
    \label{fig: pascal_examples_semseg}
\end{figure}

\begin{wraptable}[18]{r}{0.45\linewidth}
\vspace{-1em}
\caption{\small \textbf{SOTA comparison} on PASCAL.}
\label{tab: sota_pascal}
\resizebox{1.0\linewidth}{!}{
\subfloat[\centering Linear classifier.]{
    \tablestyle{4pt}{1.0}
    \label{subtab: linear_probe}
    \begin{tabular}{l|c l}
    \toprule
    \textbf{Method} & \textbf{LC} \\
    \hline
    
    \multicolumn{1}{l}{\emph{Proxy-tasks:}}\\
    \hline
    Co-Occurence~\cite{isola2015learning} & \hspace{-2.0em}13.5 & \\
    CMP~\cite{zhan2019self} & \hspace{-2.0em}16.5 & \\
    Colorization~\cite{zhang2016colorful} & \hspace{-2.0em}25.5 & \\
    \hline
    
    \multicolumn{2}{l}{\emph{Clustering:}} \\
    \hline
    IIC~\cite{xu2019invariant} & \hspace{-2.0em}28.0 & \\
    \hline
   
    \multicolumn{2}{l}{\emph{Contrastive learning:}} \\
    \hline
    Inst. Discr.~\cite{wu2018unsupervised} & \hspace{-2.0em}26.8 & \\
    MoCo~\cite{he2019momentum} & \hspace{-2.0em}45.0 & \\
    InfoMin~\cite{tian2020infomin} & \hspace{-2.0em}45.2 & \\
    SwAV~\cite{caron2020unsupervised} & \hspace{-2.0em}50.7 & \\
    \hline
    
    \multicolumn{2}{l}{\emph{Handcrafted grouping priors:}}\\
    \hline
    SegSort~\cite{hwang2019segsort}$^\dagger$ & \hspace{-2.0em}36.2 & \\
    Hierarch. Group.~\cite{zhang2020self}$^\dagger$ & \hspace{-2.0em}48.8 & \\
    MaskContrast~\cite{van2021unsupervised}  & \hspace{-2.0em}58.4 & \\
    MaskContrast~\cite{van2021unsupervised}+CRF  & \hspace{-2.0em}59.5 & \\
    \hline
    %ImageNet Class. (Supervised) & 53.1 \\
    \textbf{MaskDistill} &  \reshl{58.7}{+}{0.3} & \\
    \textbf{MaskDistill}+CRF & \reshl{62.8}{+}{3.3} & \\
    \bottomrule
    \end{tabular}}

\subfloat[\centering Clustering.]{
    \tablestyle{4pt}{1.0}
    \begin{tabular}{c l}
    \toprule
    \textbf{Clustering} \\
    
    \hline
    \\
    \hline
    
    \hspace{-2.0em}4.0 &\\
    \hspace{-2.0em}4.3 &\\
    \hspace{-2.0em}4.9 &\\
    
    \hline 
    \\
    \hline 
    
    \hspace{-2.0em}9.8 &\\
    
    \hline 
    \\
    \hline
    
    \hspace{-2.0em}4.3 &\\
    \hspace{-2.0em}3.7 &\\
    \hspace{-2.0em}4.4 &\\
    \hspace{-2.0em}4.4 &\\
    
    \hline 
    \\
    \hline
    
    \hspace{-2.0em}- &\\ \vspace{0.22em}
    \hspace{-2.0em}- &\\
    \hspace{-2.0em}35.0 &\\
    \hspace{-2.0em}- &\\
    \hline
    \reshl{45.8}{+}{10.8} & \\
    \reshl{48.9}{+}{13.9} & \\
    \bottomrule
    \end{tabular}
    \label{subtab: clustering}
}}
\end{wraptable}
Table~\ref{tab: sota_pascal} compares our results against the state-of-the-art on the PASCAL \texttt{val} set. MaskDistill consistently outperforms prior work under the linear classifier setup ($+0.3$ without CRF and $+3.3\%$ with CRF~\cite{krahenbuhl2011efficient}). Similarly, we report better results under the clustering setup ($+10.8\%$ mIoU). Figure~\ref{fig: pascal_examples_semseg} visualizes the results for our method. The model can segment semantically meaningful image regions, \eg, dogs, cars, persons, etc. In conclusion, our method learns better dense semantic representations of images than existing approaches. We further analyze our results w.r.t. different groups of works.

\textbf{(i) Proxy-tasks:} MaskDistill outperforms works that solve proxy-tasks, \eg, optical flow~\cite{zhan2019self} or colorization~\cite{zhang2016colorful}, to learn dense representations. Such proxy tasks do not capture object-level information -- an essential ingredient for tackling semantic segmentation. Differently, we capture such information explicitly by distilling object masks from self-supervised vision transformers.
\newline\textbf{(ii) Clustering:} We outperform end-to-end learning methods based on clustering, \ie, IIC~\cite{xu2015learning}. IIC is prone to degenerate solutions, as the network can easily latch onto low-level image cues like color. MaskDistill decouples feature learning and clustering to avoid this behavior.
\newline\textbf{(iii) Contrastive learning:} These works~\cite{caron2020unsupervised,he2019momentum,tian2019contrastive} learn visual representations via a contrastive loss defined at the image level. This is suboptimal because the semantic segmentation task requires disentangling the representations at the object or part level. MaskDistill achieves this via a two-step approach. 
\newline\textbf{(iv) Handcrafted grouping priors:} Finally, MaskDistill outperforms methods~\cite{hwang2019segsort,zhang2020self,van2021unsupervised} that relied on handcrafted priors to group pixels. Such priors fail to generalize to a diverse and complex dataset like PASCAL. Differently, we rely on a data-driven approach to model the pixel grouping prior. Surprisingly, we even outperform methods~\cite{hwang2019segsort,zhang2020self} ($\dagger$) that finetuned the complete ASPP decoder.

\subsubsection{Semantic Instance Segmentation}
\label{sec: sota_instseg}
\begin{figure}[t]
    \centering
    \includegraphics[width=1.0\linewidth]{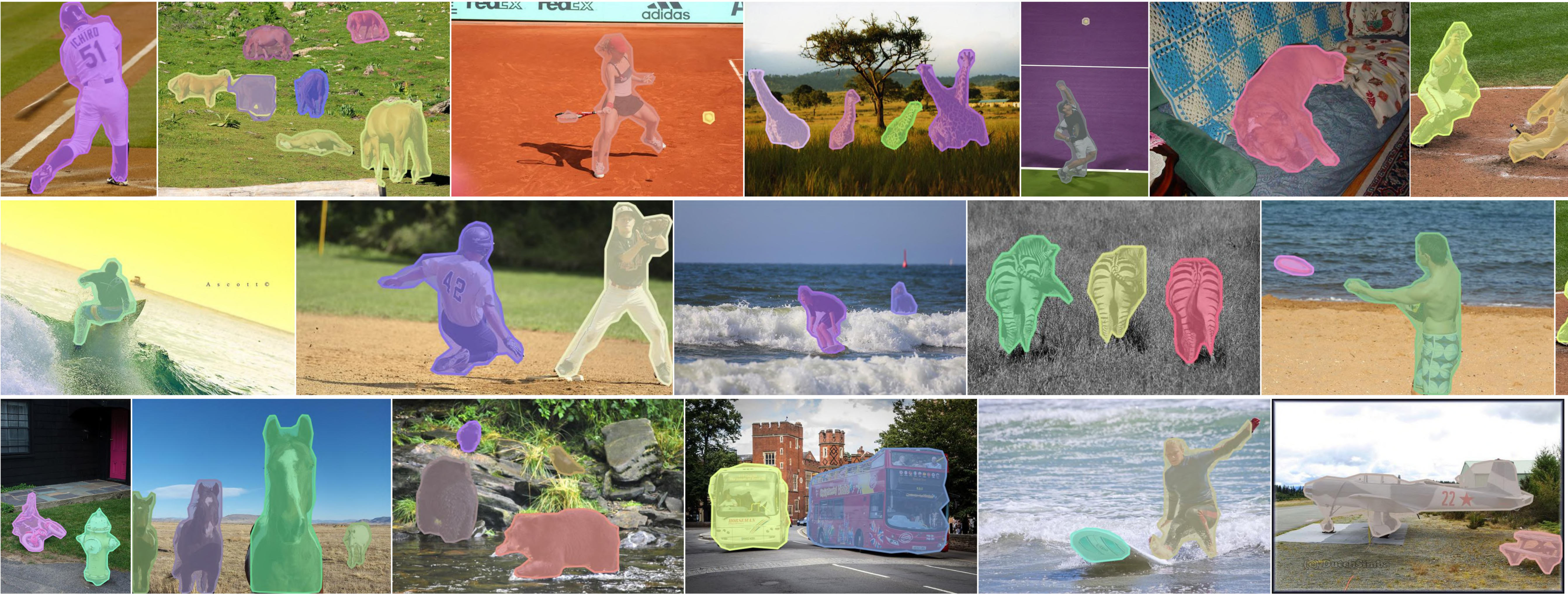}
    \caption{\textbf{Instance segmentation results} obtained with our confident object mask candidates on COCO20k.}
    \label{fig: coco20k_examples}
\end{figure}

This section evaluates our object mask candidates by comparing them against the instance segmentation ground-truth masks on PASCAL and COCO20k. We perform this analysis for (\textbf{i}) the initial mask candidates which are used as pseudo-ground-truth to train Mask R-CNN, and (\textbf{ii}) the object masks predictions from the learned Mask R-CNN model. Table~\ref{tab: instance_seg} compares our results against two other unsupervised object mask generation methods: DINO~\cite{caron2021emerging} and LOST~\cite{simeoni2021localizing}. We draw the following conclusions. First, our initial object masks outperform prior work. Our proposed mask distillation step makes fewer assumptions about the scene composition, \eg, the enclosed object area is not required to be smaller than the background~\cite{simeoni2021localizing}. MaskDistill effectively combines the advantages in prior approaches~\cite{simeoni2021localizing,caron2021emerging} to address this issue (see Section~\ref{sec: mask_distillation}). Second, the object mask candidates obtained via Mask R-CNN consistently outperform our initial object masks. We conclude that our model can better handle the multi-object setting. Figure~\ref{fig: coco20k_examples} shows several examples, where our method can retrieve multiple high-quality object masks per image.

% TABLE Instance segmentation
\begin{table}
\small
\vspace{-1.em}
\centering
\caption{\textbf{Semantic Instance Segmentation Results.} We follow the COCO-style AP$^{\text{mk}}$ metric. We consider all detections for the \texttt{multi}~\texttt{object} setting and only the most confident one for the \texttt{single}~\texttt{object} setting (see full details in the supplementary). In the class-agnostic case, the class is disregarded during evaluation. ($\dagger$) indicates that the object mask candidates from Mask R-CNN are evaluated instead of the initial object masks from the transformer~\cite{caron2021emerging} (see Section~\ref{sec: mask_distillation}).}
% -------------------------------------------
% TABLE 1: 2 subfloats
\resizebox{1.0\linewidth}{!}{
\subfloat[Class-Agnostic Instance Segmentation (\texttt{multi}) on PASCAL]{
\tablestyle{1pt}{1.0}
\begin{tabular}{x{56}|x{54}|x{54}x{54}c}
Method &
AP$^{\text{mk}}_\text{50}$ &
AP$^{\text{mk}}$ &
AP$^{\text{mk}}_\text{75}$ & \\ 
\shline
DINO~\cite{caron2021emerging} & \ressup{6.7}{} & \ressup{1.9}{} & \ressup{0.6}{} & \\
LOST~\cite{simeoni2021localizing}  & \ressup{9.6}{} & \ressup{2.9}{} & \ressup{0.9}{} & \\
\hline
MaskDistill & \reshl{12.8}{+}{3.2} & \res{4.8}{+}{1.9} & \res{0.3}{-}{0.6} & \\
MaskDistill$^\dagger$ & \reshl{24.3}{+}{14.7} & \res{9.9}{+}{7.0} & \res{6.9}{+}{6.0} & \\
\end{tabular}	
} % end of subfloat

\subfloat[Class-Agnostic Instance Segmentation (\texttt{multi}) on COCO20k]{
\tablestyle{1pt}{1.0}
\begin{tabular}{x{56}|x{54}|x{54}x{54}c}
Method &
AP$^{\text{mk}}_\text{50}$ &
AP$^{\text{mk}}$ &
AP$^{\text{mk}}_\text{75}$ & \\ 
\shline
DINO~\cite{caron2021emerging} & \ressup{1.7}{} & \ressup{0.3}{} & \ressup{0.1}{} & \\
LOST~\cite{simeoni2021localizing}  & \ressup{2.4}{} & \ressup{1.1}{} & \ressup{1.0}{} & \\
\hline
MaskDistill & \reshl{3.1}{+}{0.7} & \res{1.3}{+}{0.2} & \res{0.5}{-}{0.5} & \\
MaskDistill$^\dagger$ & \reshl{6.8}{+}{4.4} & \res{2.9}{+}{1.8} & \res{2.1}{+}{1.1} & \\
\end{tabular}	
}} % end of subfloat

% ------------------------------------------------
% TABLE 2: 2 subfloats
\small
\centering
\resizebox{1.0\linewidth}{!}{
\subfloat[Semantic Instance Segmentation (\texttt{single}) on PASCAL ]{
\tablestyle{1pt}{1.0}
\begin{tabular}{x{56}|x{54}|x{54}x{54}c}
Method &
AP$^{\text{mk}}_\text{50}$ &
AP$^{\text{mk}}$ &
AP$^{\text{mk}}_\text{75}$ & \\ 
\shline
DINO~\cite{caron2021emerging} & \ressup{13.9}{} & \ressup{4.8}{} & \ressup{2.2}{} & \\
LOST~\cite{simeoni2021localizing}  & \ressup{18.8}{} & \ressup{6.0}{} & \ressup{2.6}{} & \\
\hline
MaskDistill & \reshl{24.0}{+}{5.2} & \res{9.5}{+}{3.5} & \res{6.4}{+}{3.8} & \\
MaskDistill$^\dagger$ & \reshl{32.8}{+}{14.0} & \res{14.7}{+}{8.7} & \res{11.8}{+}{9.2} & \\
\end{tabular}	
}

\subfloat[Semantic Instance Segmentation (\texttt{single}) on COCO20k]{
\tablestyle{1pt}{1.0}
\begin{tabular}{x{56}|x{54}|x{54}x{54}c}
Method &
AP$^{\text{mk}}_\text{50}$ &
AP$^{\text{mk}}$ &
AP$^{\text{mk}}_\text{75}$ & \\ 
\shline
DINO~\cite{caron2021emerging} & \ressup{4.8}{} & \ressup{1.6}{} & \ressup{0.7}{} & \\
LOST~\cite{simeoni2021localizing}  & \ressup{8.3}{} & \ressup{2.9}{} & \ressup{1.5}{} & \\
\hline
MaskDistill & \reshl{9.9}{+}{1.6} & \res{4.1}{+}{1.2} & \res{3.1}{+}{1.6} & \\
MaskDistill$^\dagger$ & \reshl{14.6}{+}{6.3} & \res{6.6}{+}{3.7} & \res{5.5}{+}{4.0} &
\end{tabular}	

}} % end of subfloat

% ------------------------------------------------
% TABLE 3: 2 subfloats
\small
\centering
\resizebox{1.0\linewidth}{!}{
\subfloat[Semantic Instance Segmentation (\texttt{multi}) on PASCAL]{
\tablestyle{1pt}{1.0}
\begin{tabular}{x{56}|x{54}|x{54}x{54}c}
Method &
AP$^{\text{mk}}_\text{50}$ &
AP$^{\text{mk}}$ &
AP$^{\text{mk}}_\text{75}$ & \\ 
\shline
DINO~\cite{caron2021emerging} & \ressup{8.6}{} & \ressup{2.9}{} & \ressup{1.1}{} & \\
LOST~\cite{simeoni2021localizing}  & \ressup{12.1}{} & \ressup{3.8}{} & \ressup{1.5}{} & \\
\hline
MaskDistill & \reshl{15.7}{+}{3.6} & \res{6.1}{+}{2.3} & \res{4.0}{+}{2.5} & \\
MaskDistill$^\dagger$ & \reshl{24.8}{+}{12.7} & \res{10.7}{+}{6.9} & \res{8.0}{+}{6.5} & \\
\end{tabular}	
} % end of subfloat
			
\subfloat[Semantic Instance Segmentation (\texttt{multi}) on COCO20k ]{
\tablestyle{1pt}{1.0}
\begin{tabular}{x{56}|x{54}|x{54}x{54}c}
Method &
AP$^{\text{mk}}_\text{50}$ &
AP$^{\text{mk}}$ &
AP$^{\text{mk}}_\text{75}$ & \\ 
\shline
DINO~\cite{caron2021emerging} & \ressup{2.0}{} & \ressup{0.7}{} & \ressup{0.4}{} & \\
LOST~\cite{simeoni2021localizing}  & \ressup{3.3}{} & \ressup{1.2}{} & \ressup{0.6}{} & \\
\hline
MaskDistill & \reshl{4.1}{+}{0.8} & \res{1.7}{+}{0.5} & \res{1.4}{+}{0.8} & \\
MaskDistill$^\dagger$ & \reshl{7.7}{+}{4.4} & \res{3.5}{+}{2.3} & \res{2.9}{+}{2.3} & \\
\end{tabular}	
}} % end of subfloat
\label{tab: instance_seg}
\vspace{-.3em}
\end{table}
% ------------------------------------------------

\section{Discussion and Limitations}
\label{sec: limitations}
We presented a novel framework for unsupervised semantic segmentation. It first distills object masks from a self-supervised vision transformer. Next, it learns a semantic segmentation model by leveraging the most confident object mask candidates as a pixel grouping prior. This strategy addresses several limitations present in prior works. First, our method learns a pixel grouping prior in a data-driven way, rather than through handcrafted priors, which eases scaling. Second, the segmentation model does not latch onto low-level image features but learns object-level information. Third, our approach can better handle images with multiple objects. Finally, our extensive experimental evaluation shows that our method significantly outperforms the state-of-the-art.

Undoubtedly, there are still several limitations to our work. First, it's unclear how the pre-training dataset of the self-supervised vision transformer influences the quality of the object masks. Interestingly, recent research~\cite{goyal2021self,van2021revisiting} shows that we can use both object- and scene-centric datasets to learn spatially structured representations. This observation suggests that it's not crucial to train the transformer on a curated dataset, \eg, ImageNet~\cite{deng2009imagenet}. Also, there's a possibility to improve the results by scaling the pre-training dataset and model's sizes.

Another limitation of our work is that some instances can appear as a single object mask if their feature representations are strongly correlated, \eg, a motorcyclist on a motorbike. We identify several promising research directions that could potentially address this problem:
\begin{itemize}[noitemsep, topsep=0pt, left=1em..2em]
\item[--] \textbf{Multi-scale grouping:} It could be interesting to study pixel grouping priors which incorporate multi-scale features. In particular, such representations represent complementary information at the different scales~\cite{arbelaez2014multiscale,vandenhende2020mti} which could disambiguate between frequently co-occurring objects. 
\item[--] \textbf{Pre-training strategy:} We used DINO~\cite{caron2021emerging} to extract object masks. Future work could study whether better results can be obtained by changing the pre-training method, dataset, or network architecture. For example, alternative pre-training techniques~\cite{wang2020dense,desai2020virtex} could be more discriminative towards objects -- or their parts -- as they incorporate different inductive biases.
\end{itemize}
%\section*{Broader Impact}
\label{sec: broader_impact}
\paragraph{Broader Impact.} The proposed method tackles the task of semantic segmentation without using human annotations. Our evaluation shows that our method obtains promising results on several challenging benchmarks. Therefore, this research could benefit several applications where the semantic segmentation task plays an important role, \eg, medical imaging, autonomous driving, \etc. It is hard to quantify the exact societal impact at this moment. This effect will also depend on the intentions of the users and inventors. In particular, we point out that the users of our method should be aware of the different biases present in the used datasets or pre-trained models. Since our approach does not rely on carefully annotated data, such biases could potentially yield unwanted results.

\paragraph{Acknowledgment.} The authors thankfully acknowledge support by Toyota Motor Europe (TME) via the TRACE project.
% for arxiv version uncomment
%\newpage
% Define new colors
\definecolor{Highlight}{HTML}{39b54a}  % green
\definecolor{green}{HTML}{39b54a} % more transparent
\definecolor{red}{HTML}{cb4335} % red

\setcounter{section}{0}
\renewcommand\thesection{\Alph{section}}
\setcounter{figure}{0}
\setcounter{table}{0}
\renewcommand{\thefigure}{S\arabic{figure}}
\renewcommand{\thetable}{S\arabic{table}}

\begin{flushleft}
\Large{\textbf{Supplementary Materials}}
\end{flushleft}

We discuss the implementation details in Section~\ref{sec: implementation_suppl}, additional ablations in Section~\ref{sec: ablations_suppl}, additional results in Section~\ref{sec: results_suppl} and specific failure cases in Section~\ref{sec: failure_cases}.

%%%%%% Implementation Details
\section{Implementation Details}
\label{sec: implementation_suppl}
This section provides additional implementation details. The code and pre-trained models will be made available upon acceptance. All Mask R-CNN experiments (see Section~\ref{sec: mask_distillation}) were run on 4 32GB V100 GPUs. The refinement step (see Section~\ref{sec: sem_seg}) was run on 2 11GB 1080Ti GPUs. The total training time is around 20 hours. Our approach is implemented with Pytorch~\cite{pytorch}. 

\subsection{Mask R-CNN}
We follow He~\textit{et.~al.}~\cite{he2019momentum,chen2021empirical} to generate object mask candidates. In particular, we train Mask R-CNN~\cite{he2017mask} with a ResNet-50-C4 backbone while using the \texttt{Detectron2} framework~\cite{wu2019detectron2}. The model is initialized via self-supervised pre-training on ImageNet~\cite{deng2009imagenet}, \ie, MoCo~\cite{chen2021empirical}. The weights of the first two backbone stages are frozen to speedup training. Furthermore, we pick a random value from the interval [480, 800] to resize the smallest image side during training, while the image scale is 800 during inference. The learning rate is set at $0.02$ and reduced with a factor $10$ after $20$k and $22$k iterations. The model is trained for a total duration of $24$k iterations and learning rate warmup is applied for the first 100 iterations. We refer to~\cite{he2017mask,he2019momentum} for additional details.

\subsection{Linear Probing}
\label{sec: linear_probing}
During linear probing, we train a $1 \times 1$ convolutional layer on top of the frozen features. This layer is trained for $45$ epochs with a batch size of $24$. We use the SGD optimizer with weight decay $10^{-4}$ and momentum $0.9$ to update the model weights. The initial learning is set to $0.1$ and decreased with a factor of $10$ after $25$ epochs. We didn't observe improvements when training longer.

\subsection{Semantic Segmentation}
We follow the training and evaluation setup by Van Gansbeke~\etal~\cite{van2021unsupervised}. The model is DeepLab-v3~\cite{chen2017rethinking} with ResNet50~\cite{he2016deep} backbone. The model weights are updated using SGD with momentum $0.9$
and weight decay $10^{-4}$. The initial learning rate is $2\cdot 10^{-3}$ and reduced to $2\cdot 10^{-4}$ after 40 epochs of training. The total training duration is $45$ epochs with a batch size of $16$. We also apply the same \texttt{RandomHorizontalFlip} and \texttt{ScaleNRotate} augmentations during training. The original resolution is used for testing. Finally, the Hungarian algorithm~\cite{kuhn1955hungarian} matches the predicted clusters with the ground truth classes as in~\cite{van2020scan,van2021unsupervised,xu2019invariant}. The mean intersection over union (mIoU) is used as the evaluation metric.

\subsection{Semantic Instance Segmentation}
For PASCAL~\cite{everingham2010pascal}, we evaluate on the official \texttt{VOC2012} object segmentation set (2913 images). Both the \texttt{VOC2007} and \texttt{VOC2012} sets are used during training, following~\cite{he2019momentum}.
For COCO~\cite{lin2014microsoft}, we evaluate on COCO20k by following prior work~\cite{simeoni2021localizing,vo2020towards}. 
We use the mask average precision (AP) metric from \texttt{Detectron2}~\cite{wu2019detectron2} to evaluate the predictions and we report the average over 5 different runs.
We consider two scenarios during evaluation: the \texttt{multi}~\texttt{object} and \texttt{single}~\texttt{object} setting. In the \texttt{multi}~\texttt{object} setting, the model must predict all the ground truth masks. In the \texttt{single}~\texttt{object} setup, we only keep the mask with the highest confidence score for each image and select the ground truth object with the largest (bounding box) IoU for evaluation. Again, we apply the Hungarian algorithm~\cite{kuhn1955hungarian} to match the predicted clusters with the ground truth classes.
To compare with prior work, we use the publicly available code. In DINO~\cite{caron2021emerging}, we sum the attention heads and set the threshold to $0.75$. In LOST~\cite{simeoni2021localizing}, we take $400$ patch proposals. These modifications improve their performances.

\begin{figure}[H]
    \centering
    \includegraphics[width=1.0\linewidth]{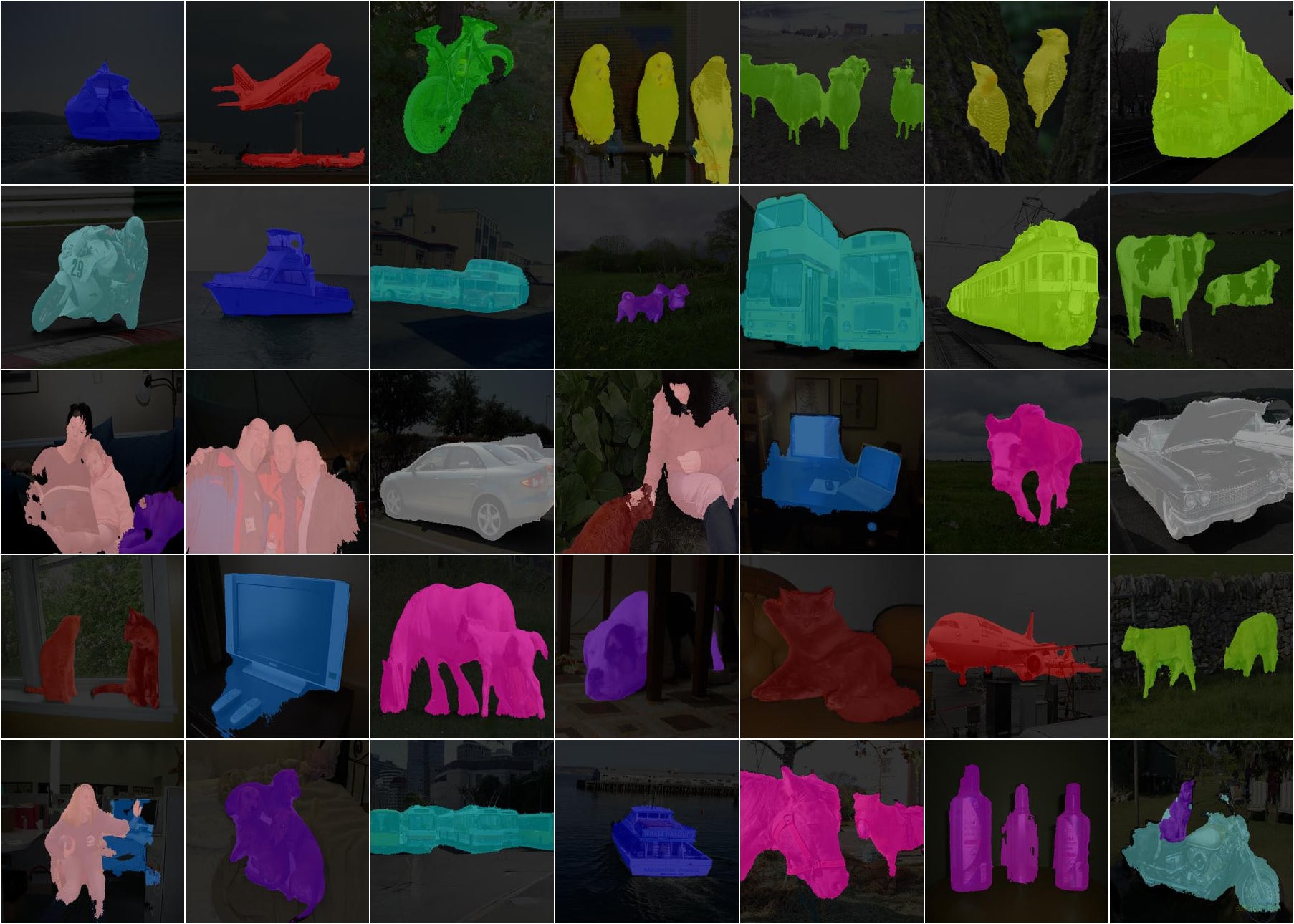}
    \caption{\textbf{Semantic segmentation results} of our method obtained under the clustering setup on PASCAL.}
    \label{fig: pascal_examples_semseg2}
\end{figure}
\begin{figure}[H]
    \centering
    \includegraphics[width=1.0\linewidth,height=8.5cm]{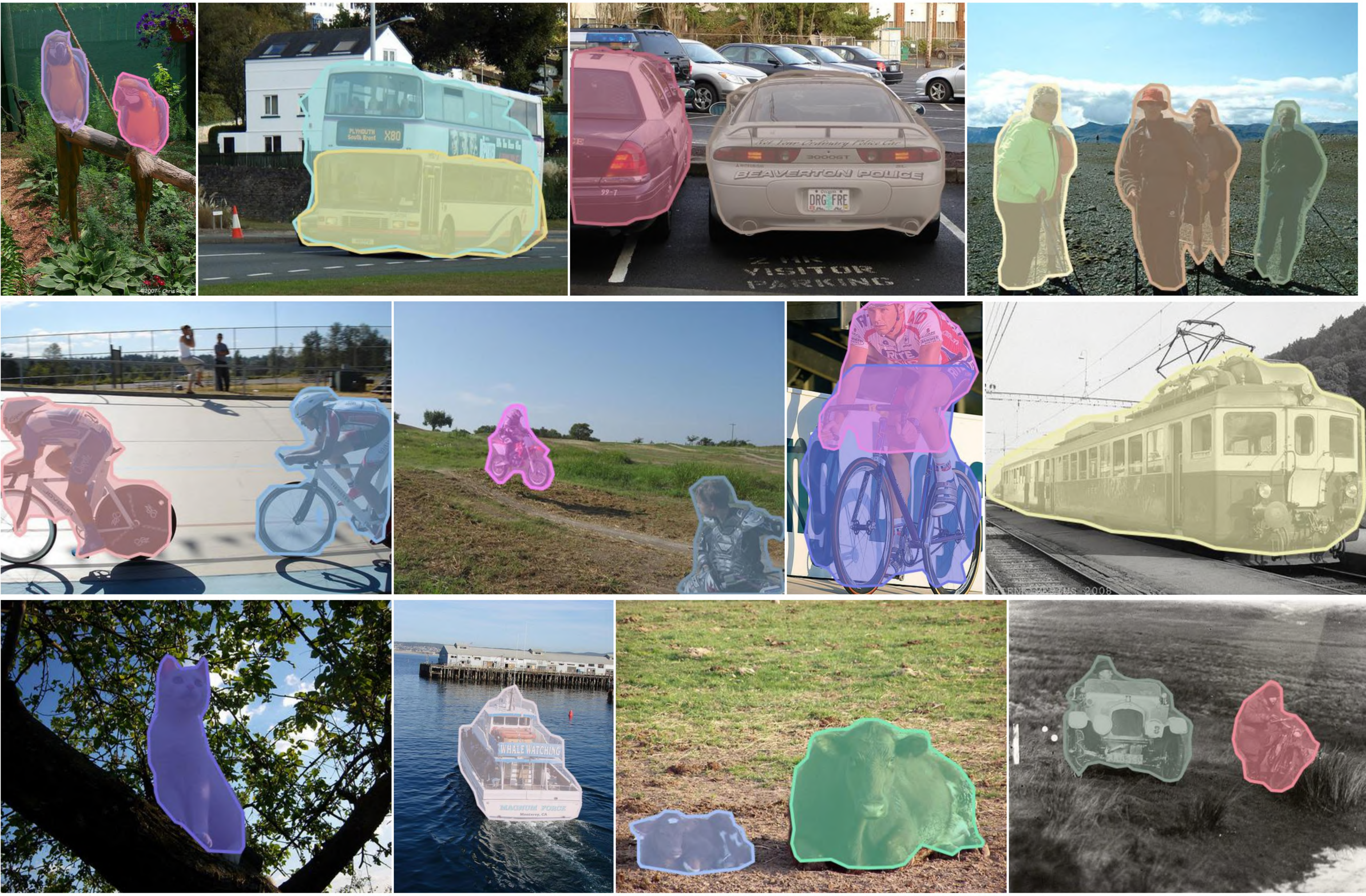}
    \caption{\textbf{Instance segmentation results} obtained with our confident object mask candidates on PASCAL.}
    \label{fig: pascal_examples}
    \vspace{-0.5em}
\end{figure}

\section{Additional Ablations}
\label{sec: ablations_suppl}
Table~\ref{tab: ablation_suppl} complements the component analysis in the main paper (see Table~\ref{tab: ablation_baselines}). We explore the predictions of the Mask R-CNN model by additionally using its confident bounding box predictions. We set the threshold $\tau$ to $0.9$ as in the main paper. Unsurprisingly the performance drops when using the bounding box predictions instead of the mask predictions from Mask R-CNN ($31.8\%$ vs. $42.0\%$).
Applying GrabCut~\cite{rother2004grabcut} to the predicted bounding box improves the results ($38.2\%$ vs. $31.8\%$).
\begin{wraptable}[13]{r}{0.45\linewidth}
    \tablestyle{4pt}{1.0}
    \caption{\textbf{Component analysis.}}
    \resizebox{1.0\linewidth}{!}{
    \begin{tabular}{l|cc}
    \toprule
    \textbf{Setup} & \textbf{\texttt{val} mIoU} \\
    \hline
    self.sup. vision transformer & 39.0 \\
    $+$ Mask R-CNN (bbox) & 31.8 \\
    $+$ GrabCut &  38.2 \\
    \hline
    self.sup. vision transformer & 39.0 \\
    $+$ Mask R-CNN (mask) & 42.0 \\
    $+$ Segmentation model & 45.8 \\
    $+$ CRF &  48.9 \\
    \bottomrule
    \end{tabular}
    }
    \label{tab: ablation_suppl}
\end{wraptable}However, it still underperforms the initial object masks from the vision transformer ($38.2\%$ vs. $39.0\%$). This supports the claim that our masks capture high-level object information, which is hard to mimic by relying on handcrafted priors as used in GrabCut. 
Finally, we point out that multiple CRF~\cite{krahenbuhl2011efficient} iterations produce additional gains ($48.9\%$ mIoU vs $45.8\%$), primarily for detailed structures. However, be aware that in order to set the importance weights of the kernels correctly, a small annotated validation set is ideal. Albeit not a required component of our framework, we conclude that iteratively updating the pseudo-ground-truth with a CRF and the model weights $\theta$ improves the segmentation results. 

\section{Additional Results}
\label{sec: results_suppl}
This section discusses additional qualitative and quantitative results on the PASCAL dataset in Section~\ref{sec: pascal_suppl} and on the COCO dataset in Section~\ref{sec: coco_suppl}.

\subsection{PASCAL}
\label{sec: pascal_suppl}
We visualize additional examples from the PASCAL dataset. In particular, Figure~\ref{fig: pascal_examples_semseg2} displays the learned clusters from our semantic segmentation model $\Phi_{\theta}$ and Figure~\ref{fig: pascal_examples} shows the confident object mask candidates. Again, we conclude that our approach discovers objects that are semantically meaningful without the necessity for annotations.

\begin{figure}[t]
    \centering
    \includegraphics[width=0.93\linewidth]{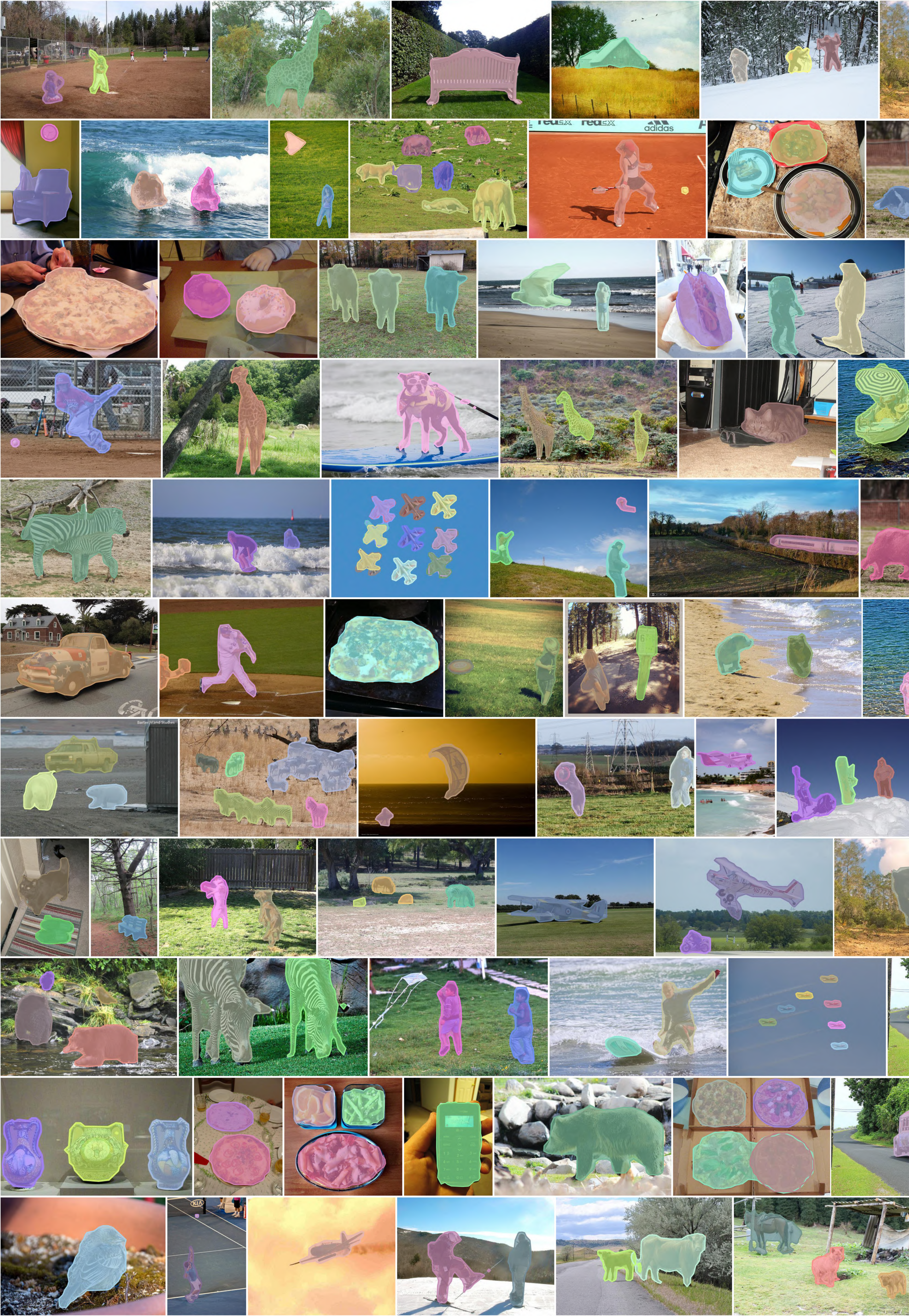}
    \caption{\textbf{Instance segmentation results} obtained with our confident object mask candidates on COCO20k.}
    \label{fig: coco_examples}
\end{figure}

Table~\ref{tab: pascal_class_iou} presents the IoU score per class. We compare with prior SOTA~\cite{van2021unsupervised} and observe large improvements for all classes.
MaskDistill discovers clusters such as \textit{bird}, \textit{cat} and \textit{train}. Not surprisingly, less discriminative classes, like \textit{chair}, \textit{table} or \textit{plant}, are more difficult to segment. Interestingly, when we apply a linear probe, the features quickly adapt to the semantics of the dataset (\textit{i.e.,} the PASCAL classes). We conclude that the model has learned semantically meaningful pixel-embeddings for different object categories. 
\begin{table}[t]
    \caption{\textbf{Semantic Segmentation Results}. We evaluate on the PASCAL \textit{val} set. ($\dagger$) indicates that we use a linear probe (see Section~\ref{sec: linear_probing}).}
    \label{tab: pascal_class_iou}
    \vspace{0.4em}
    \Huge
    \resizebox{\textwidth}{!}{
    \begin{tabular}{l|ccccccccccccccccccccc|c}
    	Method & backg. & aero & bike & bird & boat & bottle & bus & car & cat & chair & cow & table & dog & horse & mbike & person & plant & sheep & sofa & train & tv & mIoU\\
    	\shline
     	MaskContrast~\cite{van2021unsupervised} & 84.4& 68.1& 23.7& 62.6& 35.7& 0.0& 72.8& 63.0& 46.8& 0.0& 0.0& 8.5& 30.6& 28.9& 49.4& 19.4& 5.6& 34.8& 17.2& 55.7& 27.3& 35.0\\
     	\hline
     	 \textbf{MaskDistill} & 84.4& 74.7& 27.9& 70.9& 47.5& 0.0& 72.8& 33.2& 72.4& 0.0& 70.0& 29.6& 38.1& 67.5& 58.1& 28.1& 9.2& 65.8& 20.4& 65.5& 27.7& 45.8 \\
     	 \textbf{MaskDistill}+CRF & 85.4& 80.3& 28.8& 74.7& 50.4& 0.0& 72.5& 52.1& 75.7& 0.0& 76.5& 28.6& 38.7& 71.3& 63.8& 32.0& 11.2& 67.0& 20.5& 67.7& 28.7& 48.9 \\
     	\textbf{MaskDistill}$^\dagger$ & 88.1 & 80.5 & 30.9 & 76.7 & 58.2 & 52.2 & 75.7 & 70.1 & 82.7 & 12.9 & 73.3 & 35.4 & 78.8 & 72.2 & 62.1 & 52.4 & 27.9 & 73.0 & 19.7 & 70.5 & 39.0 & 58.7 \\
     	\textbf{MaskDistill}$^\dagger$+CRF & 89.8 & 83.1 & 34.0 & 85.9 & 63.3 & 45.0 & 79.1 & 70.1 & 86.3 & 16.9 & 81.5 & 38.1 & 84.0 & 74.9 & 69.7 & 63.0 & 31.3 & 78.3 & 23.3 & 74.4 & 46.1 & 62.8
    \end{tabular}}
    \label{tab: class_iou}
\end{table}

\subsection{COCO}
\label{sec: coco_suppl}
We show additional qualitative results. Figure~\ref{fig: coco_examples} displays examples of the confident object mask candidates. In contrast to PASCAL, COCO contains more complex (\ie, scene-centric) images. While the predictions are not perfect, MaskDistill detects and segments various objects fairly accurate.

\newpage
\section{Failure Cases}
\label{sec: failure_cases}
\begin{figure}[H]
    \centering
    \includegraphics[width=1.0\linewidth]{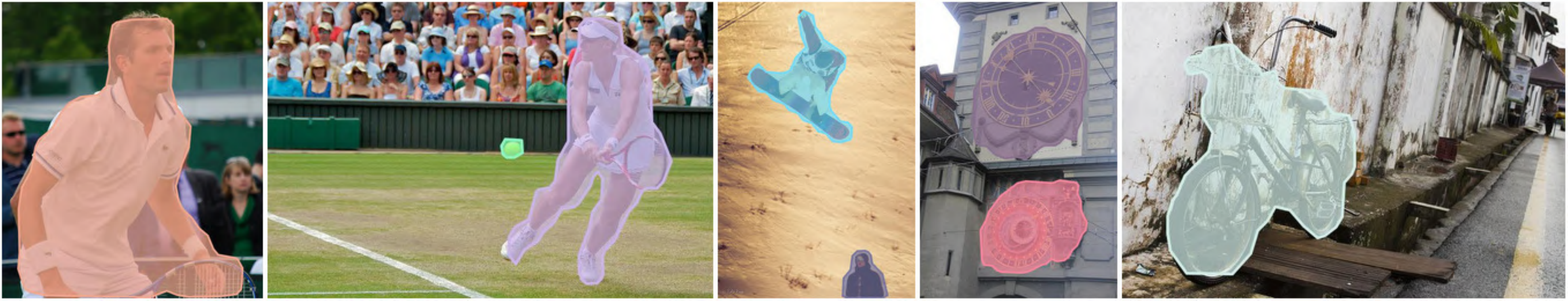}
    \caption{\textbf{Failure Cases} on COCO20k.}
    \label{fig: failure_cases}
\end{figure}

Figure~\ref{fig: failure_cases} presents several failure cases. These can be grouped as follows: 
\begin{itemize}[noitemsep, topsep=0pt, left=1.0em..2em]
\item[--] \textbf{Merging objects:} In some cases, the predicted mask encompasses multiple objects, \textit{e.g.}, "person" and "racket", "person" and "snowboard" \textit{etc.} 
\item[--] \textbf{Missing objects or parts:} The mask excludes certain objects or parts, \eg, "bike handlebars". Similarly, our model is unable to detect certain background objects, \eg, "tennis spectators".
\item[--] \textbf{Out-of-taxonomy:} The model generates object mask candidates that do not belong to the human-defined object categories, \textit{e.g.,} no class "clock" in the COCO (\textit{things) classes}. 
\end{itemize}

%\newpage
{\small
\bibliographystyle{splncs04}
\bibliography{egbib}
}

%\newpage
%\input{sections/checklist}

\end{document}